\documentclass[conference]{IEEEtran}
\IEEEoverridecommandlockouts

\usepackage{cite}
\usepackage{subcaption}
\usepackage{booktabs}
\usepackage{multirow}
\usepackage{pifont}
\usepackage{threeparttable}
\usepackage[utf8]{inputenc}
\usepackage[most]{tcolorbox}
\tcbuselibrary{listingsutf8}

\usepackage{amsmath,amssymb,amsfonts}
\usepackage{algorithmic}
\usepackage{graphicx}
\usepackage{textcomp}
\usepackage{xcolor}
\def\BibTeX{{\rm B\kern-.05em{\sc i\kern-.025em b}\kern-.08em
    T\kern-.1667em\lower.7ex\hbox{E}\kern-.125emX}}

\begin{document}

\title{OpenAVS: Open-vocabulary Audio-Visual Segmentation with Message Relay Agents
\thanks{This research is supported by the RIE2025 Career Development Fund (Award No. C233312009), administered by A*STAR, and funded by the NUS Artificial Intelligence Institute (NAII) Seed Grant (No. NAII-SG-2025-027).}
}

\author{
    \IEEEauthorblockN{
        Shengkai Chen\IEEEauthorrefmark{1}, 
        Yifang Yin\IEEEauthorrefmark{1}, 
        Jinming Cao\IEEEauthorrefmark{2}, 
        Shili Xiang\IEEEauthorrefmark{1}, 
        Zhenguang Liu\IEEEauthorrefmark{3}, and 
        Roger Zimmermann\IEEEauthorrefmark{2}
    }
    \IEEEauthorblockA{
        \IEEEauthorrefmark{1}Institute for Infocomm Research (I$^2$R), A*STAR, Singapore~  \{Chen\_Shengkai, Yin\_Yifang, Xiang\_Shili\}@a-star.edu.sg
    }
    \IEEEauthorblockA{
        \IEEEauthorrefmark{2}National University of Singapore, Singapore~ jinming.ccao@gmail.com, rogerz@comp.nus.edu.sg
    }
    \IEEEauthorblockA{
        \IEEEauthorrefmark{3}Zhejiang University, Hangzhou, Zhejiang, China~ 
        liuzhenguang2008@gmail.com
    }
}

\maketitle

\begin{abstract}
Audio-visual segmentation (AVS) aims to separate sounding objects from videos by predicting pixel-level masks based on audio signals. However, in real-world scenarios, the lack of labeled data presents harder challenges.
Existing methods primarily focus on closed-set scenarios and direct audio-visual alignment, which limits their generalization ability.
In this paper, we propose OpenAVS, a novel training-free multi-agent framework that reformulates Open-vocabulary AVS as a textual reasoning loop that effectively and flexibly aligns multi-modal signals via natural language message relaying. 
Our method seamlessly connects foundation model based specialized autonomous agents and handles AVS through three main stages:
1) Perception: \emph{Multimedia Agents} receive audio and visual streams and convert raw signals into textual descriptions.
2) Understanding: \emph{Translation Agent} receives textual multimedia descriptions, performs abstraction and digestion, aligns multi-modal representations within the textual semantic space, and then translates them into instruction prompts for segmentation. 
3) Execution: \emph{Segmentation Agent} processes these instruction prompts to generate pixel-level masks.
The \emph{Translation Agent} is a key component of our framework, enabling effective knowledge transfer from general-purpose foundation models to AVS through task-specific prompt design. Moreover, it reduces noise the textual descriptions generated by the \emph{Multimedia Agents} by enforcing prompt-wise, model-wise, and temporal consistency.
Comprehensive experiments on four benchmark datasets demonstrate the superior performance of our method. It surpasses existing methods by 4.95\% in mIoU and 2.99\% in F-score on the AVSBench dataset, and a significant gain of 12.8\% in mIoU on the zero-shot V3 dataset.
\end{abstract}


\section{Introduction}

Audio-Visual Segmentation (AVS) predicts dense masks of sounding objects in each video frame based on the audio signal. Despite the recent proliferation of deep AVS models, achieving state-of-the-art performance typically necessitates supervised training on a large-scale, fully-annotated dataset \cite{gao2023avsegformer,wang2023prompting,Mo_2024_CVPR}.
However, most of the models tailored to this task lack generalizability, reducing their effectiveness in real-world applications where domain shifts are common.

To reduce the annotation endeavor, weakly-supervised and 
unsupervised AVS methods have been proposed to alleviate the need for ground-truth labels. For instance, Point-Prompt \cite{Yu_2023_BMVC} leveraged AudioCLIP to promote the Segment Anything Model (SAM) for mask generation. OWOD-BIND \cite{bhosale2023} utilizes Open World Object Detector (OWOD) to generate class-agnostic object proposals, and links them with acoustic cues using ImageBind through cosine similarities of shared latent space embeddings. MoCA \cite{bhosale2024} adopts a self-supervised contrastive learning framework that uses DINO to extract visual embeddings and create positive and negative pairs.

\begin{figure*}[!t]
    \centering
    \includegraphics[width=0.95\linewidth]{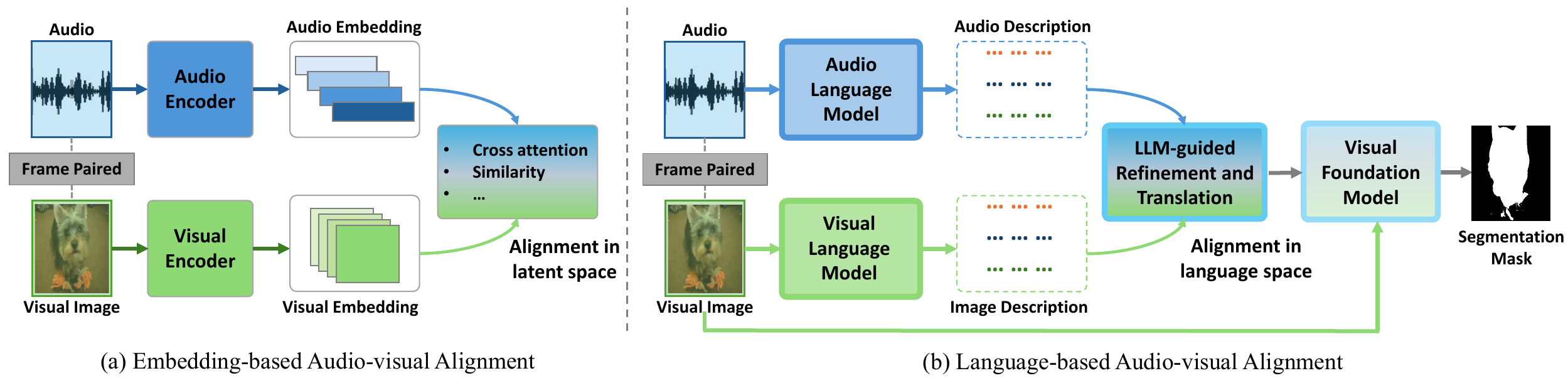}
    \caption{Different strategies for audio-visual alignment. (a) Embedding-based methods model audio-visual correlations in a latent space. (b) Language-based methods provide semantic-level alignment, enabling more effective knowledge transfer from text-audio/visual foundation models.}
    \label{fig:ava}
\end{figure*}
We observe that existing unsupervised AVS methods primarily on exploring visual foundation models, aligning audio and visual modalities within latent space based on embedding as in Figure~\ref{fig:ava}a.
These methods typically fuse audio and visual features using cross-attention \cite{bhosale2024,liu2024annotation} or similarity measures \cite{li2022languagedriven,zhou2022maskclip,girdhar2023imagebind}, often requiring fine-tuning to achieve better alignment. However, these methods face significant challenges in complex scenarios \cite{zhou2023audiovisualsegmentationsemantics}, particularly when multiple sounding objects are present in the corresponding frame. This indicates that achieving direct alignment between audio and visual modalities is challenging due to 1) limitations related to data scarcity and imbalance and 2) diverse acoustic and visual appearances of the same semantic class. 

Recent advances in Audio Language Model (ALM) and Visual Language Model (VLM) enable robust understanding and reasoning for acoustic and visual signals via natural language~\cite{zhao2023clap,liu2023llava}.
Large Language Model (LLM), which excels at interpreting textual descriptions
can generate effective prompts for Visual Foundation Model (VFM)~\cite{kirillov2023segany,liu2023grounding} to perform segmentation when given clear instructions.
Consequently, such ALMs and VLMs generally demonstrate stronger generalization compared to traditional audio-visual foundation models. 
Motivated by the Agentic AI system~\cite{acharya2025}, which cooperatively integrate foundation models and external tools to solve complex real-world problems (\emph{e.g.}, disaster relief, trading, consultant), we propose a language-based audio-visual alignment approach as depicted in Figure~\ref{fig:ava}b, to narrow the semantic gap between audio and visual content.
As illustrated in Figure~\ref{fig:framework}, we propose OpenAVS, a training-free multi-agent framework, which approaches open-vocabulary AVS task via a Perception-Understanding-Execution loop coordinated by specialized agents.
During the understanding stage, we design task-specific prompts to reduce error propagation and more importantly to facilitate effective knowledge transfer from general-purpose foundation models to AVS tasks.
We further propose to mitigate capability gaps among descriptions by enforcing 1) \emph{Model-wise Consistency}, 2) \emph{Prompt-wise Consistency}, and 3) \emph{Temporal Consistency}.
Here we summarize our contributions as follows:

\begin{itemize}
\item To the best of our knowledge, we are the first to explore language-based alignment for AVS tasks, it outperforms embedding-based methods in complex scenarios.
\item The proposed agentic framework OpenAVS, is flexible, cost-efficient, and can be effortlessly extended to support evolving multimodal reasoning modules.
\item State-of-the-art results on S4, MS3, AVSS, and V3 demonstrate that our inference-only method achieves superior reliability and generalization without training.

\end{itemize}

\section{Related Works}
Audio-visual segmentation focuses on identifying the visual regions in a frame corresponding to the sound by generating dense pixel-level predictions. 
The AVSBench dataset \cite{zhou2022avs,zhou2023audiovisualsegmentationsemantics} has been widely used for training and evaluating AVS models, covering scenarios such as single sound source, multiple sound sources, and semantic segmentation. 
Recent expansions of AVSBench include the V3 dataset~\cite{wang2023prompting} for few-shot benchmarking and the AVS-Synthetic~\cite{liu2024annotation}, which uses synthetic data to avoid manual annotations.

State-of-the-art AVS performance typically rely on the supervised training, 
a significant number of advanced fusion techniques have been proposed to learn the correlations between the audio and visual modalities~\cite{gao2023avsegformer,Mo_2024_CVPR}.
To reduce annotation cost, weakly-supervised and unsupervised AVS methods have also been explored to enhance model generalizability. For example, WS-AVS~\cite{mo2023weaklysupervised} employs weak supervision, using only class labels for class-level contrastive learning. Similarly, MoCA~\cite{bhosale2024} uses contrastive learning with foundational models in an unsupervised way.
Direct inference methods like AT-GDINO-SAM, SAM-BIND, and OWOD-BIND integrate pre-trained models such as AST \cite{gong21b_interspeech}, ImageBIND \cite{girdhar2023imagebind}, and SAM \cite{kirillov2023segany}. 
However, existing unsupervised methods are hard to generalize and require fine-tuning to obtain satisfactory results.

With the rise of foundational models, more powerful alternatives have emerged to bridge the gap between different modalities. For example, SAM offers zero-shot segmentation and GroundingDINO~\cite{liu2023grounding} performs open-set object detector with extra inputs. In addition to the aforementioned VFMs, several ALMs (Pengi~\cite{deshmukh2023pengi}, Audio Flamingo~\cite{audio_flamingo}, Qwen-Audio~\cite{qwen_audio}), VLMs (LLaVA~\cite{liu2023llava}, Qwen-VL~\cite{Qwen-VL}), and Multi-modal Language Models (MLLM) such as Qwen-Omni~\cite{qwen_omni}, MiniCPM-o-2.6~\cite{yao2024minicpm}, Omnivinci~\cite{ye2025omnivinci}, have been proposed recently, but they have not yet been successfully adapted for AVS tasks.

\section{Methodology}
\subsection{Problem Formulation}
Given paired audio signal $\mathbf{a}_i$ and visual signal $\mathbf{v}_i$ from a video source, the training-free open-vocabulary AVS aims to construct a function $\texttt{OpenAVS}_{\theta^*}$ to infer audio-visual segmentation directly,
\begin{equation}
    \mathbf{M}_i = \texttt{OpenAVS}_{\boldsymbol{\theta}^*}(\mathbf{a}_i , \mathbf{v}_i) 
\end{equation}
where $\mathbf{M}_i \in \mathbb{R}^{H \times W}$ denotes the pixel-level binary mask, with $\mathbf{M}_i=1$ indicating that the pixel belongs to a sounding object, and $\mathbf{M}_i=0$ indicating the background or a silent region.

\begin{figure*}[t]
    \centering
    \includegraphics[width=0.95\linewidth]{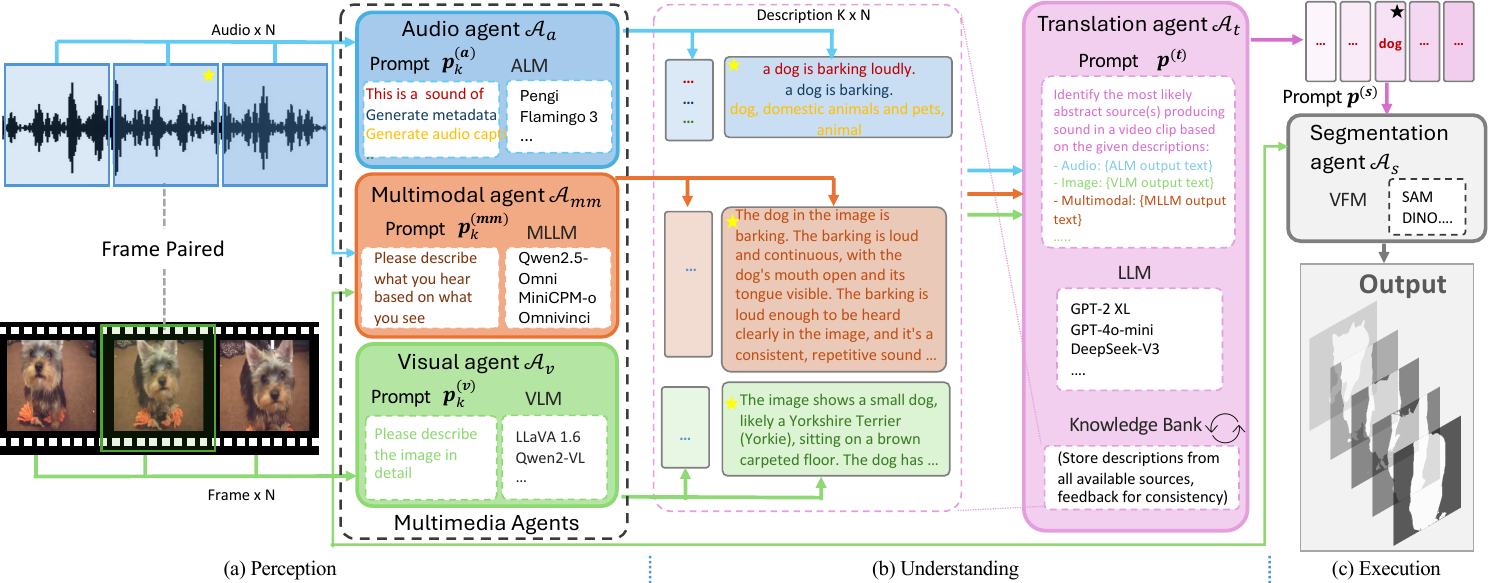}
    \caption{
    System overview of OpenAVS, three stages of reasoning  loop with default models and prompt design illustrated.
    }
    \label{fig:framework}
\end{figure*}

\subsection{Framework Overview}

We introduce OpenAVS, a training-free multi-agent system for generalized audio-visual alignment. Heterogeneous foundation model agents communicate via natural language as a unified interface. As shown in Figure~\ref{fig:framework}, a chained reasoning loop addresses AVS: (a) Perception by multimedia agents, (b) Understanding by a translation agent (maintaining a cross-modal knowledge bank), and (c) Execution by a VFM-powered segmentation agent. 




\subsection{Multi-modal Signal Perception}
The goal of the perception stage (Figure~\ref{fig:framework}a) is to map raw audio and visual signals into semantic descriptions.
Instead of direct alignment between the audio and visual embeddings, multimedia foundational models include ALM, VLM and MLLM are able to extract text from each modality, making it nature to align audio and visual semantics via text proxy.

To obtain such descriptions, we design multimedia agents with instructional prompts.
Specifically, for audio signals, \textbf{Audio Agent} ($\mathcal A_a$) receives paired audio clip to generate audio descriptions as:
\begin{equation}
\mathbf{t}^{(a)}_{i,k} = \texttt{ALM} (\mathbf{p}^{(a)}_k, \mathbf{a}_i)
\label{eq:pengi}
\end{equation}
where $\mathbf{p}^{(a)}_k$ is the audio-to-text prompt. With carefully selection (see Table~\ref{tab:ablation_gpt}), $\mathcal A_a$ can effectively describe the sounding events in the input audio. 
Similarly, for visual signals, \textbf{Visual Agent} ($\mathcal A_v$) receives paired visual image to generate visual descriptions as:
\begin{equation}
\mathbf{t}^{(v)}_{i,k} = \texttt{VLM} (\mathbf{p}^{(v)}_k, \mathbf{v}_i)
\end{equation}
where $\mathbf{p}^{(v)}_k$ is the visual-to-text prompt.
With the advance of MLLM that is able to receive different modality inputs at the same time and output a combined description, it will make the perception more convenience, such \textbf{Multi-Modal Agent} ($\mathcal A_{mm}$) usually named with ``omni'', will generate corresponding textual descriptions as:
\begin{equation}
    \mathbf{t}_{i,k}^{(mm)} = \texttt{MLLM}(\mathbf{p}^{(mm)}_k, \mathbf{a}_i, \mathbf{v}_i)
\end{equation}

\subsection{Multimedia Description Understanding}
\label{subsec:llm_guide}

Multimedia agents $\mathcal A_a$,$ \mathcal A_v$, and $\mathcal A_{mm}$ are able to extract basic semantic descriptions, however, their internal models lack segmentation specific training or fine-tuning, resulting in the unsuitable outputs for prompting VFMs, as they are prone to contain redundant or irrelevant details. 

To address these challenges, we design a \textbf{Translation Agent} ($\mathcal A_t$), who is powered by LLM to refine descriptions stored in its knowledge bank as shown in Figure~\ref{fig:framework}b. $\mathcal A_t$ translates raw Multimedia Agents' outputs into precise  segmentation prompts  
\begin{equation}
\mathbf{p}^{(s)}_i = \texttt{LLM}(\mathbf{p}^{(t)}, \mathbf{t}_i)
\end{equation}
where $\mathbf{p}^{(t)}$ is the task-specific instruction prompt, and $\mathbf t_i$ is the direct or concatenation of multi-modal descriptions ($\mathbf t_{i,k}^{(a)}, \mathbf t_{i,j}^{(v)}, \mathbf t_{i,k}^{(mm)}$).
To eliminate single source biases/mistakes and further enhance the accuracy, $\mathcal A_t$ is able to reflect internally to synthesizes robust segmentation prompts that reconcile semantic conflicts across descriptions from the same video sample with the following consistency strategies:

    \textbf{Model-wise Consistency:} Harmonize outputs from heterogeneous foundation models (ALMs, VLMs, MLLMs) to reduce potential cross-model variability;

    \textbf{Prompt-wise Consistency:} Resolves contextual discrepancies from the \emph{same} model (e.g. ALM) queried with distinct input prompts (as listed in Table~\ref{tab:ablation_gpt});
    
    \textbf{Temporal Consistency:} Enforce stability across consecutive frames by leveraging gradual audio-visual evolution within a video sample.

Theses consistencies can be represented into a unified form:
\begin{equation}
    \hat{\mathbf{p}}_i^{(s)} = \mathtt{LLM} (\mathbf p^{(c)},\dots, \mathbf t^{(j)}_{i,k},\dots)
\end{equation}
where $\mathbf p^{(c)}$ is a consistency-aware instruction prompt,  $\mathbf t_{i,k}^{(j)}$ denotes descriptions out from agent $\mathcal A_{j}, (j\in \{a, v, mm\})$ for frame $i$ under prompt variant $k$.
It can effectively remove irrelevant details and emphasize concise descriptions of the sounding object, thereby bridge modality gap to enable high-fidelity audio-visual segmentation without fine-tuning any foundation models.


\subsection{Execution: Segment Sounding Object}
Using the refined prompt ($\mathbf{p}^{(s)}_i$, $\hat{\mathbf{p}}^{(s)}_i$) from $\mathcal A_t$, the \textbf{Segmentation Agent} ($\mathcal A_s$) is able to predict segmentation masks:
\begin{equation}
\mathbf{M}_i = \texttt{VFM}(\mathbf{p}^{(s)}_i, \mathbf{v}_i)
\label{eq:vfm}
\end{equation}
As shown in Figure~\ref{fig:framework}c, $\mathcal A_s$ integrates GroundingDINO as an open-vocabulary object detector and SAM as the mask predictor.
In the experiments, we further evaluated advanced VFMs such as SAM2~\cite{ravi2024sam2} and DINO-X~\cite{ren2024dinoxunifiedvisionmodel} to investigate the performance gains without retraining any components along with the upgrade of OpenAVS.






\section{Experiments}
\newcommand{\x}{\ding{55}}
\newcommand{\y}{\ding{51}}

\subsection{Experiment Setup}
\textbf{Dataset} We evaluate our proposed method on four benchmark datasets, namely \textbf{S4}, \textbf{MS3}, \textbf{AVSS}, and \textbf{V3} from AVSBench. Following previous work~\cite{wang2023prompting,liu2024annotation}, we convert the semantic labels of \textbf{AVSS} and \textbf{V3} to object labels by $\mathbf{Y}^{(object)}_i = \min(\mathbf{Y}^{(semantic)}_i, \mathbf{1})$.

\textbf{Evaluation Metric}
We adopt the mean Intersection over Union (mIoU, $\mathcal M_J$) and F-score ($\mathcal M_F$) to evaluate our model. A higher mIoU value implies better region similarity, while a higher F-score indicates improved contour accuracy.

\textbf{Implementation Details}
We implemented three OpenAVS variants with progressively enhanced capabilities:
\begin{itemize}
\item  \textbf{OpenAVS-Lite}: Empolys only $\mathcal A_a$ using lightweight Pengi, with \emph{Prompt-wise} and \emph{Temporal} Consistencies.
\item \textbf{OpenAVS}: Enables both $\mathcal A_a$ and $\mathcal A_v$ , with both \emph{Model-wise} and \emph{Prompt-wise} Consistencies.
\item \textbf{OpenAVS-Large}: Extends \textbf{OpenAVS} with multiple ALM ensemble for $\mathcal A_a$ with \emph{Model-wise} Consistency.
\end{itemize}

Experiments were conducted on a single L40 46GB GPU machine with AMD EPYC 64-core CPU.

\begin{table}[t]
\begin{threeparttable}
\centering
\caption{Comparison with open-vocabulary unsupervised AVS methods on S4, MS3, AVSS datasets.}
\setlength{\tabcolsep}{1mm}
\label{tab:training-free}
\begin{tabular}{rcccccccccccc}
\toprule
\multirow{2}{*}{Method} & \multirow{2}{*}{TF\tnote{a}} &  \multicolumn{2}{c}{S4} & \multicolumn{2}{c}{MS3} & \multicolumn{2}{c}{AVSS-Binary} \\
& & $\mathcal M_J$ & $\mathcal M_F$ & $\mathcal M_J$ & $\mathcal M_F$ &$\mathcal M_J$ & $\mathcal M_F$  \\ 
\midrule
AT-GDINO-SAM    & \y & 0.380 & 0.460 & 0.250 & 0.290 &-&-\\
Point-Prompt    & \y & 0.403 & 0.515 & 0.288 & 0.333 &-&-\\
SAM-BIND        & \y & 0.420 & 0.510 & 0.280 & 0.360 &-&-\\
Box-Prompt      & \y & 0.512 & 0.615 & 0.418 & 0.478 &-&-\\
OWOD-BIND       & \y & 0.580 & 0.670 & 0.340 & 0.440 &-&-\\
OV-AVSS (USSL)  & \x & 0.486 & 0.616 & 0.361 & 0.427 &0.525 &0.617\\
AL-Ref-SAM2     & \y & \textbf{0.705} & \textbf{0.811} & 0.486 & 0.535 &0.592 &0.662\\ 
\midrule
OpenAVS-Lite    & \y & 0.582 & 0.689 & 0.483 & \underline{0.565} & 0.593 &0.654 \\
OpenAVS-Lite (SAM2) & \y & 0.639 & 0.725 & \underline{0.512} & \textbf{0.587} & 0.617 &0.667\\
OpenAVS (SAM2)  & \y & 0.680  & 0.764  & 0.511 & 0.541 & \underline{0.651} & \underline{0.704} \\
OpenAVS-Large (SAM2) & \y & \underline{0.701}	& \underline{0.782} & 	\textbf{0.525}	& 0.557 &\textbf{0.659} &\textbf{0.711}   \\
\bottomrule
\end{tabular}
\begin{tablenotes}
    \item[a] TF: Training-Free.
\end{tablenotes}
\end{threeparttable}
\end{table}


\begin{table}[t]
\caption{Comparison with few-shot and zero-shot AVS on V3 dataset. Our method is not trained with any `seen' shots.}
\centering
 \setlength{\tabcolsep}{0.8mm}
    \begin{tabular}{rcccccccccc}
    \toprule
    Subset  & \multirow{2}{*}{TF} & \multicolumn{2}{c}{0-shot}  & \multicolumn{2}{c}{1-shot} & \multicolumn{2}{c}{3-shot} & \multicolumn{2}{c}{5-shot}  \\
    Method & &  $\mathcal M_J$ & $\mathcal M_F$ &$\mathcal M_J$ & $\mathcal M_F$ &$\mathcal M_J$ & $\mathcal M_F$ & $\mathcal M_J$ & $\mathcal M_F$ & \\ 
    \midrule
    SAM-Fusion &\x & 0.463 & 0.630 & 0.504 & 0.671 & 0.571 & 0.719 & 0.608 & 0.741 \\
    TPAVI    &\x & 0.530 & 0.707  & 0.561 & 0.754 & 0.632 & 0.767 & 0.639 & 0.783 \\ 
    AVSegFormer &\x& 0.543 & 0.715  & 0.583 & 0.764 & 0.642 & \textbf{0.774} & 0.652 & 0.785 \\  
    GAVS     &\x   & 0.547 & 0.722  & 0.629 & \textbf{0.768} & 0.663 & \textbf{0.774} & 0.678 & \textbf{0.795} \\
    \midrule
    OpenAVS-Lite   &\y & 0.663 & 0.736   & 0.692 & 0.755 & 0.695 & 0.760& 0.696 & 0.761 \\
    (with SAM2) &\y & \textbf{0.675} & \textbf{0.741}  & \textbf{0.699} & 0.757 & \textbf{0.702} & 0.761& \textbf{0.703} & 0.762 \\
    \bottomrule
    \end{tabular}
\label{tab:v3_zero_shot}
\end{table}

\subsection{Comparison with the State-of-the-arts}
As shown in Table~\ref{tab:training-free}, we compare our proposed \mbox{OpenAVS} to the 7 unsupervised AVS methods, including Point-Prompt and Box-Prompt~\cite{Yu_2023_BMVC}, AT-GDINO-SAM, SAM-BIND, and OWOD-BIND~\cite{bhosale2023}, OV-AVSS~\cite{ov-avss}, and AL-Ref-SAM2~\cite{al_ref_sam2}.
OpenAVS-Lite is also compared with SOTA training-based methods, SAM-Fusion~\cite{wang2023prompting}, TPAVI~\cite{zhou2022avs}, AVSegFormer~\cite{gao2023avsegformer}, and GAVS~\cite{wang2023prompting} on few- and zero-shot V3 settings~\cite{wang2023prompting} in Table~\ref{tab:v3_zero_shot}.


\subsubsection{Results on open-vocabulary unsupervised AVS}
Training-free open-vocabulary AVS approaches use foundation models for open-vocabulary audio-visual segmentation directly on test samples, without data collection or model training. 
As shown in Table~\ref{tab:training-free}, OpenAVS outperforms existing SOTA methods by a significant margin on MS3 and AVSS datasets, where multiple sounding objects are present in the frames. 
Upgrading OpenAVS-Lite's backbone to SAM2 (OpenAVS-Lite (SAM2)) further boosts the performance, demonstrating the framework’s ability to evolve and improve in step with advances in foundation models.
It's worth noting that \mbox{AL-Ref-SAM2} outperforms OpenAVS-Large by 0.4\% on S4, but they utilize GPT-4o/GPT-4-turbo with a costly two-step refinement process ($\approx 16.32\times$ higher inference cost than ours; see Table~\ref{tab:model_comparison} and Supplementary Materials for more details).
Our approach maintains competitive accuracy by using only GPT-4o-mini with one single-shot text query, highlighting its cost-performance flexibility. 


\subsubsection{Results on few-shot and zero-shot AVS}

\begin{table}[t]
\centering
 \setlength{\tabcolsep}{0.7mm}
\caption{Prompt selection of $\mathcal A_a$, parameter size of ALMs, and the contribution of $\mathcal A_t$ in OpenAVS-Lite with S4.}
\begin{tabular}{p{2.9cm}|cr|cc|cc}
\toprule
 \multirow{2}{*}{Prompt $\mathbf p_k^{(a)}$}  &\multirow{2}{*}{ALM}  & \multirow{2}{*}{\#Para}& \multicolumn{2}{c|}{w/o $\mathcal A_t$ (\x)} & \multicolumn{2}{c}{with $\mathcal A_t$} (\y) \\
& & & $\mathcal M_J$ & $\mathcal M_F$ &$\mathcal M_J$ & $\mathcal M_F$   \\ \midrule
- \textit{This is a sound of} & Pengi & 0.3B & 0.549 & 0.635 &0.568 & 0.657\\  
- \textit{Generate metadata}  & Pengi &0.3B & 0.551 & 0.640 & 0.564 & 0.653\\
- \textit{Generate audio caption} & Pengi &0.3B & {0.560}& 0.650 & 0.567 & 0.656 \\
\multirow{2}{=}{- \textit{Please describe the audio in detail}} &  Flamingo 3 & 8.2B & 0.558 & {0.672} & 0.551 & 0.684 \\
 & Qwen2.5-Omni & 10.7B & 0.548 & 0.663 & 0.563 & {0.692} \\
\bottomrule
\end{tabular}
\label{tab:ablation_gpt}
\end{table}
\begin{table}[t]
 \setlength{\tabcolsep}{1.8mm}
\centering
\caption{Efficiency analysis with varying ALMs and VLMs for multimedia agents $\mathcal A_a$ and $\mathcal A_v$ on the S4 dataset.}
\label{tab:alm_ablation}
\begin{tabular}{c|c|c|ccccc}
\toprule
Variant & ALM & VLM  &  $\mathcal M_J$ & $\mathcal M_F$   \\ \midrule
OpenAVS-Lite&Pengi &- & 0.630 & 0.718   \\
OpenAVS-Lite&Pengi (PC) &- & 0.639 & 0.725  \\
OpenAVS-Lite& Audio Flamingo 3&- & 0.592 & 0.684  \\
OpenAVS-Lite&Qwen2.5-Omni &- & 0.604 & 0.695 \\ 
\midrule
OpenAVS&Pengi  & Qwen2.5-Omni & 0.678 & 0.762 \\
OpenAVS&Pengi (PC) & Qwen2.5-Omni & 0.680 & 0.764  \\
OpenAVS&Pengi (PC) & Qwen3-VL & 0.681 & 0.765  \\
OpenAVS&Qwen2.5-Omni & Qwen2.5-Omni & 0.665 & 0.752  \\
\midrule
OpenAVS-Large &All ALMs & Qwen2.5-Omni & 0.701 & 0.782 \\
\bottomrule
\end{tabular}
\end{table}
As shown in Table \ref{tab:v3_zero_shot}, our training-free OpenAVS-Lite significantly outperforms existing approaches in mIoU, with absolute improvements of 12.8\% in zero-shot setting, 7.0\%, 3.9\%, and 2.5\% in 1-shot, 3-shot, and 5-shot scenarios.
It is worth noting that OpenAVS-Lite performs inference exclusively on the test split even for few-shot datasets and does not use any seen samples for training.
Additionally, our method achieves the highest F-score in the zero-shot setting, whereas all competing methods are trained using `seen' data from V3.
These results verify the robustness of our proposed method to unseen scenarios.

\subsection{Ablation Study}




\subsubsection{Prompt and ALM selection for $\mathcal A_a$}
We evaluate the sensitivity of using 3 distinct prompts $\mathbf p_{i,k}^{(a)}$ listed in its original work~\cite{deshmukh2023pengi} for Pengi in $\mathcal A_a$. The results in Table~\ref{tab:ablation_gpt} indicate that using \textit{``Generate audio caption''} to prompt Pengi is generally more robust for AVS tasks.
Then, we compared model sizes of ALMs for $\mathcal A_a$, as listed in Table~\ref{tab:ablation_gpt}.
While larger ALMs like Flamingo and Qwen achieve performance gains, these improvements are not proportional to the growth in their parameter sizes. Since these models are designed for more general audio tasks, such as speech processing, the sound-focused Pengi is more task-aligned with AVS.

\subsubsection{Mode Analysis of Multimedia Agents}

\begin{table}[t]
    \caption{Comparison with Multi-modal LLM on S4.}
    \centering
     \setlength{\tabcolsep}{1.2mm}
    \begin{tabular}{ccccccccc}
    \toprule
    \multirow{2}{*}{Model} & \multirow{2}{*}{Mode} & \multicolumn{2}{c}{S4} & \multicolumn{2}{c}{MS3} & \multicolumn{2}{c}{AVSS} \\
    & & $\mathcal M_J$ & $\mathcal M_F$ &$\mathcal M_J$ & $\mathcal M_F$ &$\mathcal M_J$ & $\mathcal M_F$ \\
    \midrule
    OpenAVS&  $\mathcal A_{mm}$ & 0.662 & 0.752 & 0.417 & 0.470 & 0.624 & 0.685  \\
    (Qwen2.5-Omni)& $\mathcal A_a + \mathcal A_v$ & 0.665 &0.752 & {0.514} & {0.540} & {0.642} & {0.693}  \\
    \midrule
    OpenAVS& $\mathcal A_{mm}$ & 0.655 & 0.746 & 0.426 & 0.471 & 0.624 & 0.682 \\
    (MiniCPM-o-2.6)& $\mathcal A_a + \mathcal A_v$  & {0.665} &	{0.752} &	0.472 &	0.503 &	0.639& 	0.691  \\
    \midrule
    OpenAVS &$\mathcal A_{mm}$  & {0.665} & 	0.747 & 0.429 & 0.457 &	0.635 &	0.687  \\
    (Omnivinci)& $\mathcal A_a + \mathcal A_v$  & 0.656 & 0.743 & 0.456 & 0.491 & 	0.628 & 0.681 \\
    \midrule
    OpenAVS-Large & $\mathcal A_a + \mathcal A_v$ & 0.701 & 0.782 & 0.525 & 0.557 & 0.659 & 0.711 \\
    \bottomrule
    \end{tabular}
    \label{tab:mm_direct_or_not}
\end{table}


With $\mathcal A_t$ and $\mathcal A_s$ fixed, Table~\ref{tab:alm_ablation} evaluates OpenAVS variants on S4 dataset.
Recall that OpenAVS-Lite only enables $\mathcal A_a$, OpenAVS enables both $\mathcal A_a$ and $\mathcal A_v$, and OpenAVS-Large ensembles multiple ALMs for $\mathcal A_t$.
Enabling $\mathcal A_v$ increases the mIoU ($\mathcal M_J$) by 4.9\%, and the Qwen3-VL empowered $A_v$ achieves the highest mIoU for OpenAVS, as it extracts more informative messages for downstream agents.
However, $\mathcal A_v$ requires more time to generate descriptions, as images typically contain richer information than audio, leading to longer encoding processes and more output tokens.
Moreover, we report the results obtained by multimodal LLMs, denoted as $\mathcal{A}_{mm}$, which are capable of processing audio and visual inputs simultaneously, in Table~\ref{tab:mm_direct_or_not}.
Generally, the $\mathcal A_a + \mathcal A_v$ mode tends to yield better or competitive results compared to its corresponding $\mathcal{A}_{mm}$ mode, which is observed across both datasets and models.
One possible reason is that most MLLMs are trained for general tasks other than AVS. Comparatively, our framework employs AVS-specific prompts to bridge $\mathcal{A}_a$ and $\mathcal{A}_v$ in the separate mode, thereby improving MLLM performance on AVS tasks.

\subsubsection{Performance–Cost analysis for $\mathcal A_t$}
\begin{table}[t]
\centering
 \setlength{\tabcolsep}{0.5mm}
\caption{Performance and cost comparison on S4.}
\begin{tabular}{lllllll}
\toprule
Model & $\mathcal A_t$ & $\mathcal A_s$ & $\mathcal M_J$ & $\mathcal M_F$ & Cost/video \\
\midrule
 OpenAVS-Lite & GPT-2 XL & GDINO+SAM & 0.431 & 0.561 & \$0 \\
 OpenAVS-Lite &DeepSeek-V3 & GDINO+SAM  & 0.576 & 0.681 & 0.00125 CNY \\
 OpenAVS-Lite &GPT-4o-mini & GDINO+SAM  & 0.582 & 0.689 & \$0.000154 \\
 OpenAVS-Lite &GPT-4o-mini& GDINO+SAM2 & 0.638 & 0.728 & \$0.000154 \\
 OpenAVS & GPT-4o-mini& GDINO+SAM2 & 0.680  & 0.764 & \$0.00135  \\
 OpenAVS-Large &GPT-4o-mini& GDINO+SAM2 & 0.701  & 0.782  & \$0.00163 \\
 \multicolumn{3}{c}{AL-Ref-SAM2~\cite{al_ref_sam2}} & 0.705 & 0.811& \$0.0266 \tiny ($\times$16.3)\\
\bottomrule
\end{tabular}
\label{tab:model_comparison}
\end{table}

As discussed above, $\mathcal{A}_t$ is essential for adapting existing MLLMs effectively to AVS tasks by designing task-specific prompts that bridge $\mathcal{A}_a$ and $\mathcal{A}_v$.
Here we compare the performance and cost with varying LLMs and report the results in Table~\ref{tab:model_comparison}. We also report the estimated cost of AL-Ref-SAM2~\cite{al_ref_sam2}, computed using 50 randomly selected samples from S4, as a baseline. For OpenAVS-Lite, the transition from GPT-2 XL to DeepSeek-V3 and GPT-4o-mini leads to significant improvements in both mIoU and F-score, with only a marginal increase in cost. Notably, GPT-4o-mini paired with GDINO+SAM2 achieves the best performance while keeping the cost per video as low as \$0.000154, demonstrating that high-quality segmentation can be achieved without relying on expensive models. For both OpenAVS and OpenAVS-Large, the image descriptions generated by the VLM are substantially longer than the audio descriptions produced by the ALM. As a result, the number of prompt tokens processed by the LLM increases significantly, leading to improved performance at the expense of higher per-sample cost for both OpenAVS and OpenAVS-Large.
These experiments show that $\mathcal A_t$ not only improves the sounding object descriptions, but also aligns the outputs of the ALM and VLM to enhance performance.
With a larger budget, more advanced models such as GPT-4-turbo (as used in AL-Ref-SAM2) can be adopted for further improvement.

\subsubsection{Discussion on the Limitations of VFMs for $\mathcal{A}_s$}
\begin{figure}[t]
\centering
    \begin{subfigure}[b]{0.32\linewidth}
        \centering
        \includegraphics[height=1.4cm]{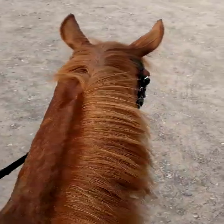}\includegraphics[height=1.4cm]{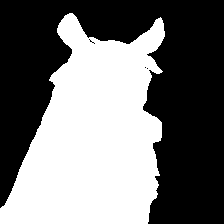}
        \caption{\scriptsize Ground Truth}
        \label{fig:figure1}
    \end{subfigure}
    \begin{subfigure}[b]{0.32\linewidth}
        \centering
        \includegraphics[height=1.4cm]{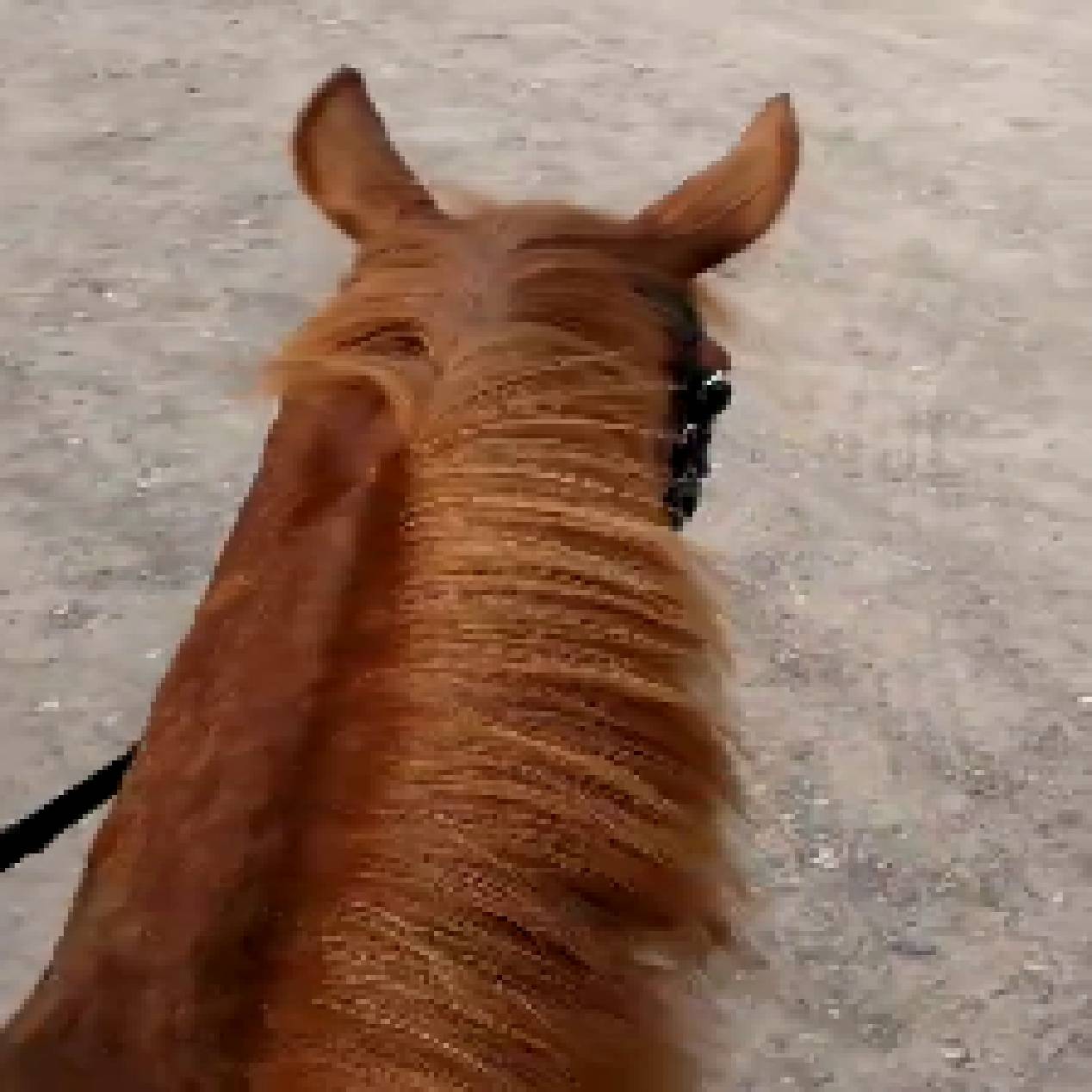}\includegraphics[height=1.4cm]{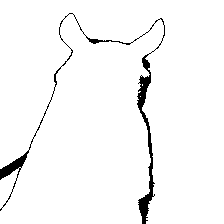}
        \caption{\scriptsize DINO 1.0}
    \end{subfigure}
    \begin{subfigure}[b]{0.32\linewidth}
        \centering
        \includegraphics[height=1.4cm]{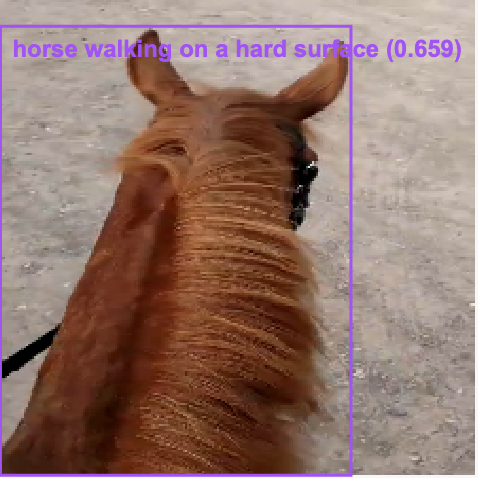}\includegraphics[height=1.4cm]{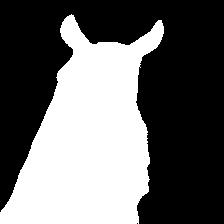}
        \caption{\scriptsize DINO-X}
        \label{fig:figure2}
    \end{subfigure}
    \caption{GDINO: \textit{``horse walking on a hard surface''}. 
    }
    \label{fig:3x2grid}
\label{fig:limit}
\end{figure}

Although OpenAVS can be constrained by the choice of foundation models, its performance is expected to improve as these models continue to advance.
For example, as shown in Figure~\ref{fig:limit}, the current GroundingDINO model (DINO 1.0) fails to ground the box in a case where the image presents an uncommon view of a horse, making it a challenging recognition task. The upgraded commercial version, DINO-X, achieves remarkable performance improvements, highlighting its strong potential for real-world applications.
Furthermore, upgrading from SAM to SAM2 for $\mathcal A_s$ also leads to substantial improvements, as shown in Tables~\ref{tab:training-free} and \ref{tab:v3_zero_shot}.

\subsection{Visualizations}
\newcommand{\imgw}{1.6cm}
\begin{figure*}[t]
    \centering
    \footnotesize
    \begin{threeparttable}
    \begin{tabular}
    {p{1.9cm}p{1.9cm}p{1.9cm}p{1.9cm}p{1.9cm}p{1.9cm}p{1.9cm}}
    \toprule
    \textbf{Raw Image} & \textbf{Ground Truth}  & \textbf{AVSS (USSL)} & \textbf{OV-AVSS} & \textbf{OpenAVS (CL)}\tnote{a} & \textbf{OpenAVS (PO)}\tnote{b} & \textbf{OpenAVS} \\
    \midrule
    \includegraphics[width=\imgw]{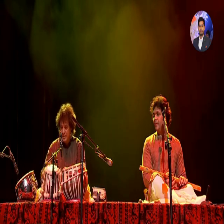} & \includegraphics[width=\imgw]{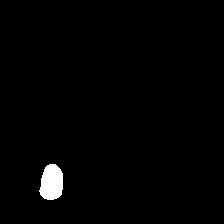} & \includegraphics[width=\imgw]{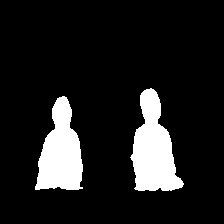} & \includegraphics[width=\imgw]{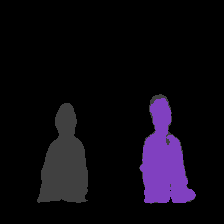} & \includegraphics[width=\imgw]{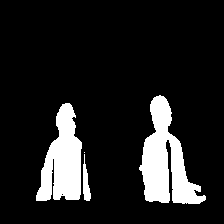} & \includegraphics[width=\imgw]{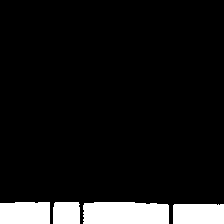} & \includegraphics[width=\imgw]{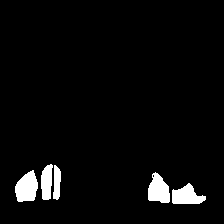} \\
    \multicolumn{7}{r}{\textbf{Clip-level class label}: tabla \quad \textbf{Pengi}: a drum loop is being played. \quad \textbf{OpenAVS}: drum}\\
    \midrule
    \includegraphics[width=\imgw]{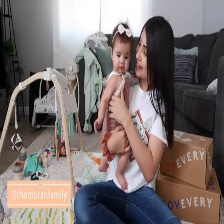} & \includegraphics[width=\imgw]{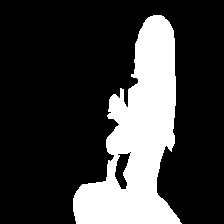} & \includegraphics[width=\imgw]{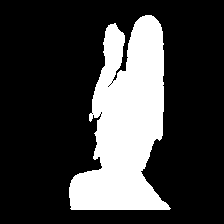} & \includegraphics[width=\imgw]{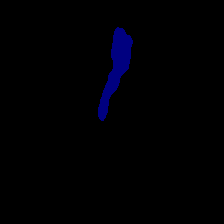} & \includegraphics[width=\imgw]{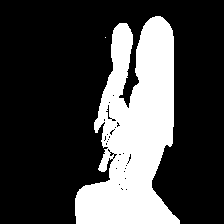} & \includegraphics[width=\imgw]{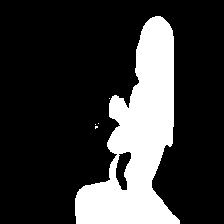} & \includegraphics[width=\imgw]{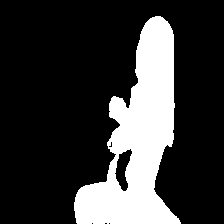} \\
    \multicolumn{7}{r}{\scriptsize \textbf{Clip-level class label}: baby\_woman \quad \textbf{Pengi}: a woman is saying something and a man is saying something. \quad \textbf{OpenAVS}: a woman}\\
    \bottomrule
    \end{tabular}
    \begin{tablenotes}
    \item[a] Use Clip-level class label text \quad $^\text{b}$ Use raw Pengi output text
    \end{tablenotes}
    \caption{Visual comparison of OpenAVS variants vs. baselines on challenging AVS cases.}
\label{tab:peformance-subset}
\end{threeparttable}
\end{figure*}
Figure~\ref{tab:peformance-subset} illustrates the qualitative comparison between OpenAVS and other approaches.
As OV-AVSS is primarily designed for semantic segmentation task, we include its semantic masks for reference.
OpenAVS shows strong robustness, even in challenging scenarios where objects of the same category (\emph{e.g.}, a baby and a woman) appear in the same frame.
The results also highlight that OpenAVS can handle a variety of text prompt inputs, which significantly influence the performance.
Using Pengi outputs directly (OpenAVS (PO)) exhibits instability in producing masks, as the raw outputs are not well-suited for $\mathcal A_s$. This underscores the importance and effectiveness of our $\mathcal A_t$, which bridges the modality gap between foundational models. 



\section{Conclusion}

We device a novel training-free agentic framework \mbox{OpenAVS} to generate pixel-level masks of the sounding objects in each video frame. This framework demonstrates a promising direction for using language as a bridge to integrate diverse task-oriented foundation models. By leveraging richer contextual information conveyed through text-based message passing among multimodal agents, the translation agent can more effectively reason about appropriate prompts for the segmentation agents. This reasoning-driven framework shows strong potential for extension to more complex and challenging domains.






\bibliographystyle{IEEEtran}
\bibliography{IEEEabrv,ref}

\newpage
\appendix
\renewcommand{\thefigure}{A\arabic{figure}}
\renewcommand{\thetable}{A\arabic{table}}
\setcounter{figure}{0}
\setcounter{table}{0}


\section{LLM Cost Analysis}
\begin{table}[ht]
\centering
\caption{OpenAI LLM API costs per 1 million tokens}
\label{tab:llm_costs}
\begin{tabular}{l c c}
\toprule
\textbf{Model} & \textbf{Input Cost (\$/1M)} & \textbf{Output Cost (\$/1M)} \\
\midrule
GPT-4o-mini    & 0.15 & 0.60 \\
GPT-4o         & 2.50 & 10.00 \\
GPT-4-turbo    & 10.00 & 30.00 \\
\bottomrule
\end{tabular}
\end{table}

Table~\ref{tab:llm_costs} collected related LLM costs\footnote{https://platform.openai.com/docs/pricing} realted to OpenAVS and AL-Ref-SAM2.
AL-Ref-SAM2 incurs higher cost per sample than OpenAVS, due to it require 1 stage to call GPT-4o and 1 stage to call GPT-4-turbo, while OpenAVS only calls GPT-4o-mini in the understanding stage once.

\section{Datasets and Evaluation Metrics}

\subsection{Dataset details}

We evaluate our method on the following benchmark datasets that vary in complexity, scale, and semantic diversity.

\begin{itemize}
    \item \textbf{S4} consists of 4,932 video clips covering 23 distinct classes, each containing a single sounding object. This dataset serves as a standard benchmark for evaluating segmentation performance in clean and well-separated audio-visual scenarios.
    \item \textbf{MS3} contains 424 samples from the same 23 classes as S4 but features multiple concurrent sounding sources per clip, making it more challenging due to overlapping audio events and increased ambiguity in visual cues.
    \item \textbf{AVSS (V2)} extends the \textbf{S4} and \textbf{MS3} datasets scale and diversity, featuring 12,356 videos across 70 categories. It includes upgraded versions of the original 5,356 videos and 7,000 newly collected multi-source clips.
    \item \textbf{V3} includes 11,356 video clips spanning 70 object categories, formed by merging \textbf{MS3} with the \textbf{V2} dataset~\cite{zhou2023audiovisualsegmentationsemantics}. It introduces a larger and more diverse label space and is split into seen and unseen categories to facilitate evaluation under few-shot and zero-shot settings~\cite{wang2023prompting}. This makes V3 particularly suitable for assessing the open-vocabulary generalization capabilities of audio-visual segmentation methods.
\end{itemize}

Together, these datasets provide a comprehensive testbed for evaluating performance across both standard and open-set scenarios, covering a range of challenges from single-source to multi-sources audio-visual events and from limited to large-scale category diversity.

\subsection{Evaluation metrics}

The evaluation metrics used in this paper are defined as follows.  

\textbf{Mean of Intersection over Union (mIoU).}  
For binary audio-visual segmentation task, the IoU for the foreground class is given by:
\begin{equation}
    IoU = \frac{TP}{TP + FP + FN}
\end{equation}
where $TP$, $FP$, and $FN$ denote true positives, false positives, and false negatives, respectively.  

The mean IoU (mIoU) is then defined as:
\begin{equation}
    \mathcal M_J = \frac{1}{2}\left(IoU_{\text{fg}} + IoU_{\text{bg}}\right)
\end{equation}
where $IoU_{\text{fg}}$ corresponds to the segmented object (foreground), and $IoU_{\text{bg}}$ corresponds to the background.  

\textbf{F-score.}  
The generalized $\mathcal F_\beta$ score is defined as:
\begin{equation}
    \mathcal M_F = \frac{(1+\beta^2) \times \text{Precision}\,\times\,\text{Recall}}
    {\beta^2 \times \text{Precision} + \text{Recall}}
\end{equation}
where we set $\beta^2 = 0.3$, following~\cite{zhou2022avs}.













\section{Additional Experimental Results}

\subsection{Extended Results on AVSS}

On the binarized AVSS dataset, we compare our method not only with unsupervised approaches, such as OV-AVSS~\cite{ov-avss} and AL-Ref-SAM2~\cite{al_ref_sam2} (Table~I), but also with supervised methods that leverage ground-truth masks, including Audio-SAM, SAM-Fusion, GAVS~\cite{wang2023prompting}, and TPAVI~\cite{zhou2022avs} in Table~\ref{tab:compare_avss}. The results demonstrate that our unsupervised, training-free OpenAVS not only outperforms other unsupervised competitors but also achieves performance competitive with these supervised methods.

\begin{table}[t]
\centering
\caption{Comparison with supervised and unsupervised methods on AVSS dataset. GT refers to ground-truth labels.}
\begin{tabular}{rcccc}\toprule
Method &TF & GT & $\mathcal M_J$ & $\mathcal M_F$ \\\midrule
Audio-SAM & \x  & \y &0.574 &0.684 \\
SAM-Fusion & \x  & \y &0.602 &0.724 \\
TPAVI & \x  & \y&0.625 &0.756 \\
GAVS & \x  & \y&0.677 &0.788 \\
\midrule
OV-AVSS (USSL) & \x &  \x & 0.525 & 0.617 \\
AL-Ref-SAM2 & \y  &  \x & 0.592 & 0.662 \\
OpenAVS & \y & \x & 0.593 & 0.654 \\
OpenAVS-Lite (SAM2) & \y & \x & 0.617 & 0.667 \\
OpenAVS-Large (SAM2)  & \y & \x & 0.659 & 0.711 \\
\bottomrule
\end{tabular}
\label{tab:compare_avss}
\end{table}

\subsection{Example of LLM-Guided Prompt Translation}

\begin{figure}[t]
    \begin{subfigure}[b]{0.2\linewidth}
        \centering
        \includegraphics[height=1.8cm]{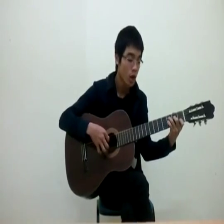}
        \caption{\scriptsize Raw frame\\ ~}
        \label{fig:raw_fig}
    \end{subfigure}
    \hfill
    \begin{subfigure}[b]{0.25\linewidth}
        \centering
        \includegraphics[height=1.8cm]{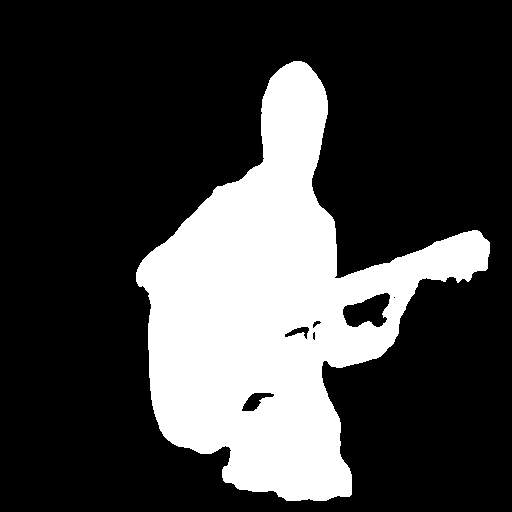}
        \caption{\scriptsize \textit{A person is playing a guitar}}
        \label{fig:pengi_direct}
    \end{subfigure}
    \hfill
    \begin{subfigure}[b]{0.2\linewidth}
        \centering
        \includegraphics[height=1.8cm]{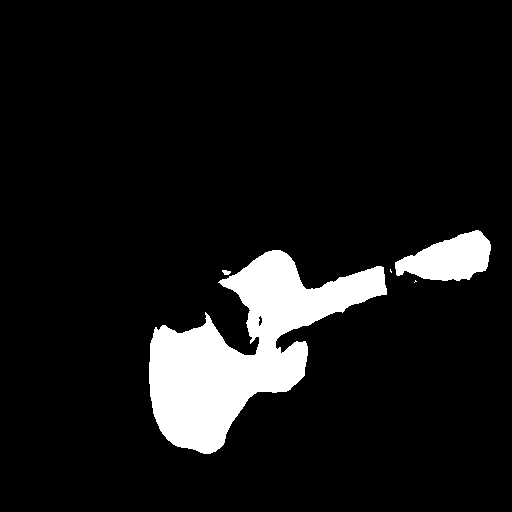}
        \caption{\scriptsize \textit{Refined output: Guitar}}
        \label{fig:gpt4_translate}
    \end{subfigure}
    \hfill
    \begin{subfigure}[b]{0.2\linewidth}
        \centering
        \includegraphics[height=1.8cm]{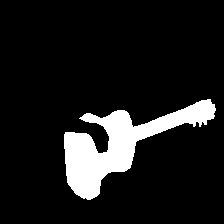}
        \caption{\scriptsize GT mask\\ ~}
        \label{fig:gt_fig}
    \end{subfigure}
    \caption{Translate audio description from (b) to (c) to improve segmentation accuracy.}
    \label{fig:llm_translator}
\end{figure}

As described in Section~III, OpenAVS has translation agent $\mathcal A_t$ to refine the output of the descriptions from multimedia agents, mitigating issues caused by misleading or ambiguous texts. For instance, as shown in Figure~\ref{fig:llm_translator}, although a person appears in the frame, he was not producing any sound. Using the raw output $\mathbf{t}^{(a)}_{i,k}$ directly could mislead the segmentation agent $\mathcal A_{s}$ into segmenting irrelevant objects since it will capture the word \textit{``person''} as well.

\begin{table}[t]
\centering
\caption{Ablation study of consistency strategies}
\label{tab:ablation_time_moe}
\begin{tabular}{cc|cc|c}
\toprule
Prompt & Frame   & \multicolumn{2}{c|}{S4} & \#API calls\\
 Consistency & Consistency & $\mathcal M_J$ & $\mathcal M_F$  & per video \\ \midrule
 \x & \x & 0.567 & 0.656 & 5 \\  
  \x & \y & 0.569 & 0.674 & 5 \\
 \y & \x & 0.581 & 0.687 & 15 \\
 \y &\y &  \textbf{0.582} & \textbf{0.689} & 15 \\
\bottomrule
\end{tabular}
\end{table}

The results in Table~III verify the effectiveness of the $\mathcal A_t$ with performance gains observed across all cases when enhanced by LLM (GPT-4o-mini). Moreover, Table~\ref{tab:ablation_time_moe} shows that both \emph{Prompt-wise} and \emph{frame-wise} (\emph{Temporal}) consistency strategies are capable of addressing aforementioned potential issues. 
While frame consistency may not yield as significant an improvement as prompt consistency, it did not require any additional LLM API calls hence is meaningful for balancing cost, inference time, and performance in real-world applications.

\subsection{Impact of Using Clip-Level Weak Labels}

We further investigate the effect of replacing the $\mathcal A_t$ outputs with clip-level weak labels, defining a ``soft'' upper bound of our OpenAVS-Lite, denoted as \mbox{OpenAVS-Lite*}. This approach is inspired by WS-AVS~\cite{mo2023weaklysupervised}, which leverages clip-level annotations instead of pixel-level masks to reduce annotation costs in AVS tasks.

As shown in Table~\ref{tab:weak_label}, using weak labels provides only marginal improvements on the S4 and MS3 datasets. However, performance still falls short of our OpenAVS (SAM2) and OpenAVS-Large (SAM2) results reported in Table~I. One contributing factor is that clip-level labels apply to the entire audio clip, including silent periods, which can lead to inaccurate segmentation since they fail to capture temporal variations in the scene.

\begin{table}[t]
\centering
 \setlength{\tabcolsep}{1.5mm}
\caption{Impact of using clip-level text labels (OpenAVS-Lite*)}
\label{tab:weak_label}
\begin{tabular}{rcccccc}
\toprule
\multirow{2}{*}{Method} & \multicolumn{2}{c}{S4} & \multicolumn{2}{c}{MS3} & \multicolumn{2}{c}{V3 (0-shot)} \\  
        &  $\mathcal M_J$ & $\mathcal M_F$ &  $\mathcal M_J$ & $\mathcal M_F$ & $\mathcal M_J$ & $\mathcal M_F$ \\ \midrule
OpenAVS-Lite  & 0.582 & 0.689 & 0.483 & 0.565 & 0.663 & 0.736 \\  
OpenAVS-Lite (SAM2) & 0.638 & 0.728 & 0.512 & 0.587 & 0.675 & 0.741 \\
OpenAVS-Lite* & 0.648 & 0.748 & 0.536 & 0.600 & 0.656 & 0.726 \\  
\bottomrule
\end{tabular}
\end{table}

\section{Additional Visual Illustrations}

\begin{figure*}
    \centering
    \begin{threeparttable}
    \scriptsize
    \begin{tabular}
    {p{1.8cm}p{1.8cm}p{1.8cm}p{1.8cm}p{1.8cm}p{1.8cm}p{1.8cm}}
    \toprule
    \textbf{Raw Image} & \textbf{Ground Truth}  & \textbf{AVSS (USSL)} & \textbf{OV-AVSS} & \textbf{OpenAVS (CL)}\tnote{a} & \textbf{OpenAVS (PO)}\tnote{b} & \textbf{OpenAVS} \\
    \midrule
    \includegraphics[width=\imgw]{figures/bhDqxWQUIXg_1_2_raw.png} & \includegraphics[width=\imgw]{figures/bhDqxWQUIXg_1_2_gt.png} & \includegraphics[width=\imgw]{figures/bhDqxWQUIXg_1_2_ov1.png} & \includegraphics[width=\imgw]{figures/bhDqxWQUIXg_1_2_ova.png} & \includegraphics[width=\imgw]{figures/bhDqxWQUIXg_1_2_clstxt.png} & \includegraphics[width=\imgw]{figures/bhDqxWQUIXg_1_2_pengi.png} & \includegraphics[width=\imgw]{figures/bhDqxWQUIXg_1_2_ours.png} \\
    \multicolumn{7}{r}{\textbf{Clip-level class label}: tabla \quad \textbf{Pengi}: a drum loop is being played. \quad \textbf{OpenAVS}: drum}\\
    \midrule
    \includegraphics[width=\imgw]{figures/EnpqqJXs0aY_2_3_raw.png} & \includegraphics[width=\imgw]{figures/EnpqqJXs0aY_2_3_gt.png} & \includegraphics[width=\imgw]{figures/EnpqqJXs0aY_2_3_ov1.png} & \includegraphics[width=\imgw]{figures/EnpqqJXs0aY_2_3_ova.png} & \includegraphics[width=\imgw]{figures/EnpqqJXs0aY_2_3_clstxt.png} & \includegraphics[width=\imgw]{figures/EnpqqJXs0aY_2_3_pengi.png} & \includegraphics[width=\imgw]{figures/EnpqqJXs0aY_2_3_ours.png} \\
    \multicolumn{7}{r}{\textbf{Clip-level class label}: baby\_woman \quad \textbf{Pengi}: a woman is saying something and a man is saying something. \quad \textbf{OpenAVS}: a woman}\\
    \midrule
    \includegraphics[width=\imgw]{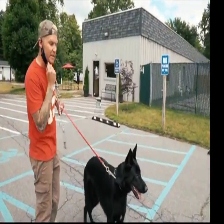} & \includegraphics[width=\imgw]{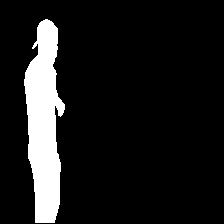} & \includegraphics[width=\imgw]{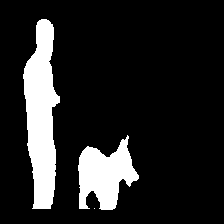} & \includegraphics[width=\imgw]{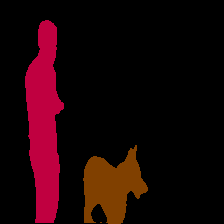} & \includegraphics[width=\imgw]{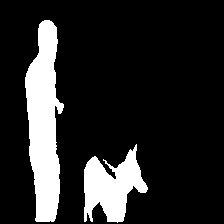} & \includegraphics[width=\imgw]{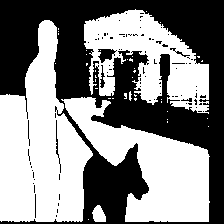} & \includegraphics[width=\imgw]{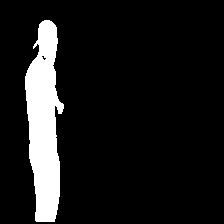} \\
    \multicolumn{7}{r}{\textbf{Clip-level class label}: man\_dog \quad \textbf{Pengi}: a man is speaking and a microphone is being used. \quad \textbf{OpenAVS}: a man}\\
    \midrule
    \includegraphics[width=\imgw]{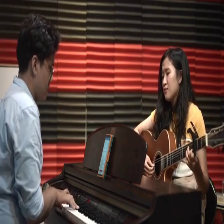} & \includegraphics[width=\imgw]{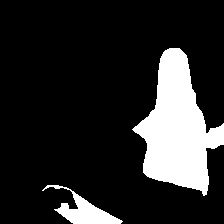} & \includegraphics[width=\imgw]{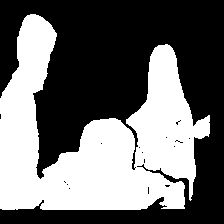} & \includegraphics[width=\imgw]{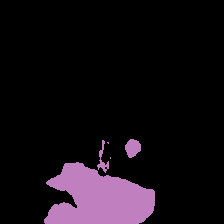} & \includegraphics[width=\imgw]{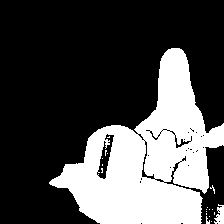} & \includegraphics[width=\imgw]{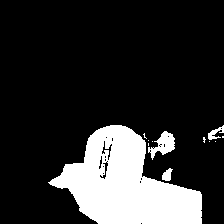} & \includegraphics[width=\imgw]{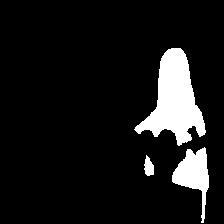} \\
    \multicolumn{7}{r}{\textbf{Clip-level class label}: woman\_piano\_guitar \quad \textbf{Pengi}: a woman is singing a song. \quad \textbf{OpenAVS}: a woman}\\
    \midrule
    \includegraphics[width=\imgw]{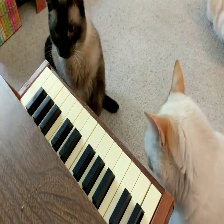} & \includegraphics[width=\imgw]{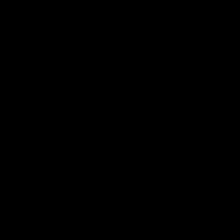} & \includegraphics[width=\imgw]{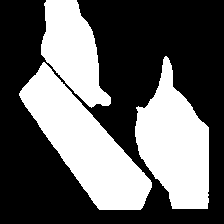} & \includegraphics[width=\imgw]{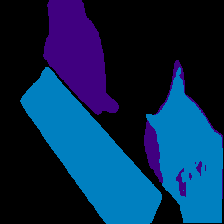} & \includegraphics[width=\imgw]{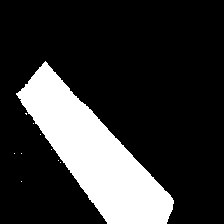} & \includegraphics[width=\imgw]{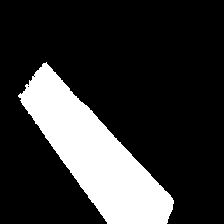} & \includegraphics[width=\imgw]{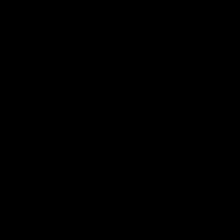} \\
    \multicolumn{7}{r}{\textbf{Clip-level class label}: piano \quad \textbf{Pengi}: a bell is ringing and a person is saying something. \quad \textbf{OpenAVS}: a bell}\\
    \midrule
    \includegraphics[width=\imgw]{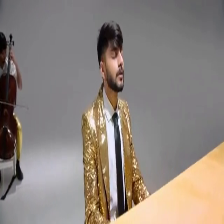} & \includegraphics[width=\imgw]{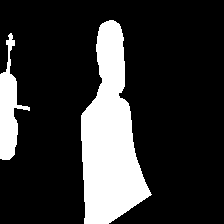} & \includegraphics[width=\imgw]{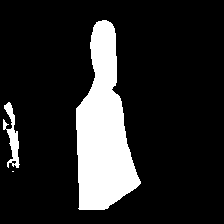} & \includegraphics[width=\imgw]{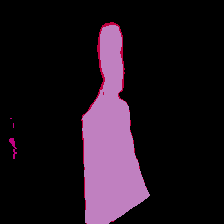} & \includegraphics[width=\imgw]{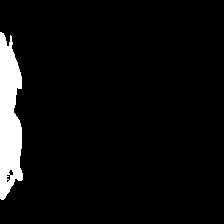} & \includegraphics[width=\imgw]{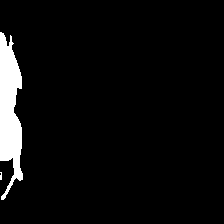} & \includegraphics[width=\imgw]{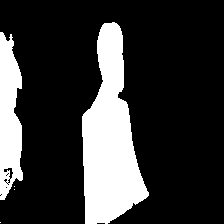} \\
    \multicolumn{7}{r}{\textbf{Clip-level class label}: violin\_man\_piano \quad \textbf{Pengi}: someone is singing a song. \quad \textbf{OpenAVS}: someone}\\
    \bottomrule
    \end{tabular}
    \begin{tablenotes}
    \item[a] Use Clip-level class label text \quad $^\text{b}$ Use raw Pengi output text
    \end{tablenotes}
\end{threeparttable}
    \caption{Visual comparison of OpenAVS variants vs. baselines on challenging AVS cases.}
    \label{fig:peformance}
\end{figure*}

To supplement the main results presented in the Figure~4, we provide additional qualitative examples in Figure~\ref{fig:peformance} to further illustrate the effectiveness and robustness of OpenAVS under challenging audio-visual conditions. These examples, omitted from the main paper due to space constraints, follow the same evaluation setup.

The visual comparisons showcase various OpenAVS variants alongside baseline methods, emphasizing cases with ambiguous sound sources, overlapping audio events, and visually complex scenes. OpenAVS consistently generates more accurate and temporally coherent segmentation masks by leveraging language-guided open-set inference. These extended results support the findings reported in the main paper and offer deeper insight into the generalization capabilities of our approach.

It is worth noting that method OV-AVSS was originally designed for semantic segmentation tasks and follows a two-stage pipeline: (a) Universal Sound Source Localization (USSL) and (b) Open-Vocabulary Classification (OVC). The first stage module USSL is trained to locate all objects in an image given the corresponding audio signal. Subsequently, the second stage module OVC leverages holistic class-level label text to semantically filter the localized objects using CLIP. Since our work does not incorporate semantic information, we compare against the first-stage module of their approach, referred to as OV-AVSS (USSL) in Table~I.

\section{Prompts and Inputs for LLM translation strategies}
\label{sec:appendix_llm_prompt}

\begin{figure*}[t]
    \centering
    \includegraphics[width=\linewidth]{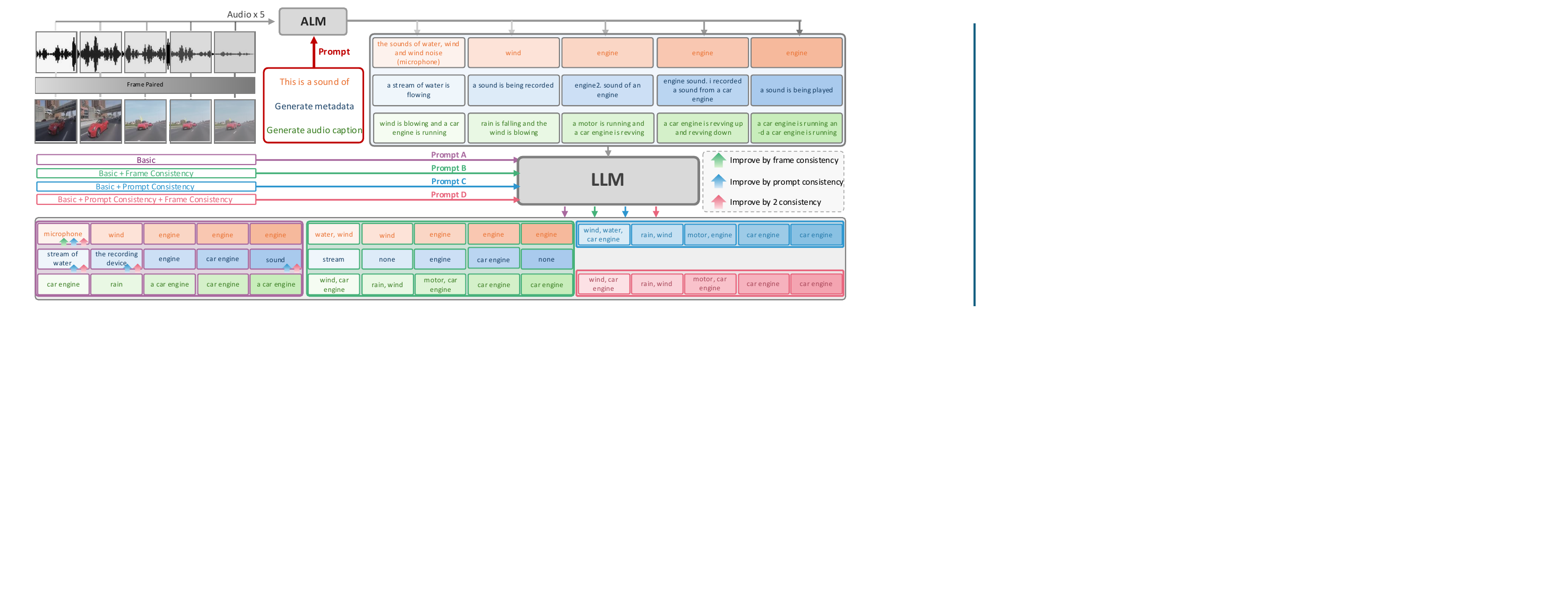}
    \vspace{-2mm}
    \caption{Illustration of Prompt-wise and Frame-wise Consistency in the $\mathcal A_t$ for Improving $\mathcal A_a$ Outputs.}
    \label{fig:2consistency}
    \vspace{-2mm}
\end{figure*}

As described in Section~III, we designed 3 consistency strategies, namely \emph{Prompt-wise}, \emph{Frame-wise} (\emph{Temporal}), and \emph{Model-wise} consistency, to enhance the translation quality produced by the $\mathcal M_t$. An illustration of these strategies is shown in Figure~\ref{fig:2consistency}.

For each 1-second video segment, its audio is fed into the $\mathcal A_a$, which takes a fixed text prompt and generates a description of what it ``hears'' in that segment. In our setup, we use three fixed prompts: \textit{``This is a sound of''}, \textit{``Generate metadata''}, and \textit{``Generate audio caption''}, as shown in Figure~\ref{fig:2consistency}.
Additionally, we use general prompt \textit{``Please describe the audio in detail''} for other ALMs like Audio Flamingo and Qwen2.5-Omni, and \textit{``Please describe the image in detail''} for VLM like Qwen2.5-Omni.
These prompts can be interpreted as simulating different expert perspectives on the same audio input. Each prompt produces one output per segment (or frame), so a 5-second video results in 5 frames of output per prompt.


The details of consistency prompting and the format of user input are presented in the following sections.

\subsection{Basic translator}

\tcbset{
  boxstyle1/.style={
    colback=purple!5,
    colframe=purple!70!black,
    arc=6pt,
    boxrule=0.5pt,
    fontupper=\ttfamily,
    enhanced,
    breakable,
    left=10pt, right=10pt, top=10pt, bottom=10pt,
    width=\linewidth,
  },
  boxstyle2/.style={
    colback=orange!5,
    colframe=orange!50!black,
    arc=6pt,
    boxrule=0.5pt,
    fontupper=\ttfamily,
    enhanced,
    breakable,
    left=10pt, right=10pt, top=10pt, bottom=10pt,
    width=\linewidth,
  }
}

\textbf{System prompt}:

\begin{tcolorbox}[boxstyle1, title=System Prompt for basic Translator]
You are participating in a competitive game where your goal is to identify the most likely abstract source(s) (e.g., human, instrumental, etc.) that is/are producing sound in a given audio clip. This clip was broken down into several frames, each containing multiple audio outputs generated by different AIs, representing sounds at a specific timestamp.

Each frame corresponds to a different moment in the same video clip and some frames may contain no sound-producing objects at all, or the text output could provide misleading information.

~

Your task:

- Identify and output only the object(s) producing sound in each frame.

- For each frame, provide your guess in one line, (seperate by comma if multiple objects), enclosed in with <answer> and </answer> tag pair.
\end{tcolorbox}

\noindent\textbf{User input}:

\begin{tcolorbox}[boxstyle2, title=User Input for Basic Translator]

<frame0>

~     wind is blowing and a car engine is running

</frame0>

<frame1>

 ~    rain is falling and the wind is blowing

</frame1>

<frame2>

  ~   a motor is running and a car engine is revving.

</frame2>

<frame3>

   ~  a car engine is revving up and revving down.

</frame3>

<frame4>

    ~ a car engine is running and a car engine is running.

</frame4>
\end{tcolorbox}

\subsection{Translator with \textcolor{cyan}{Prompt}-wise consistency}

\textbf{System prompt}:

\begin{tcolorbox}[boxstyle1, title=System Prompt for Translator with Prompt Consistency]
    
You are participating in a competitive game where your 
goal is to identify the most likely abstract source(s)
(e.g., human, instrumental, etc.) that is/are producing
sound in a given audio clip. This clip was broken down
into several frames, each containing multiple audio 
outputs generated by different AIs, representing sounds
at a specific timestamp.

Each frame corresponds to a different moment in the same
video clip and some frames may contain no sound-producing
objects at all, or the text output could provide 
misleading information.

~

Your task:

\textbf{\textcolor{cyan}{- Analyze the outputs from all audio AIs in each frame together.}}

- Identify and output only the object(s) producing sound
  in each frame.

- For each frame, provide your guess in one line, 
  (seperate by comma if multiple objects), enclosed in 
  with <answer> and </answer> tag pair.
\end{tcolorbox}

\noindent\textbf{User input}:

\begin{tcolorbox}[boxstyle2, title=User Input for Translator with Prompt Consistency]
<frame0>

~ <exp1>wind is blowing and a car engine is running </exp1>
 
~ <exp2>a stream of water is flowing. </exp2>
 
~ <exp3>the sounds of water, wind and wind noise (microphone) </exp3>

</frame0>

<frame1>

 ~<exp1>rain is falling and the wind is blowing. </exp1>

~ <exp2>a sound is being recorded. </exp2>

 ~<exp3>wind </exp3>

</frame1>

<frame2>

 ~<exp1>a motor is running and a car engine is revving. </exp1>

 ~<exp2>engine2. sound of an engine </exp2>

 ~<exp3>engine </exp3>

</frame2>

<frame3>

 ~<exp1>a car engine is revving up and revving down. </exp1>

 ~<exp2>engine sound. i recorded a sound from a car engine. </exp2>

 ~<exp3>engine </exp3>

</frame3>

<frame4>

~ <exp1>a car engine is running and a car engine is running. </exp1>

~ <exp2>a sound is being played. </exp2>

~ <exp3>engine </exp3>

</frame4>
\end{tcolorbox}

\subsection{Translator with \textcolor{teal}{Frame}-wise consistency}

\textbf{System prompt}:

\begin{tcolorbox}[boxstyle1, title=System Prompt for Translator with Frame Consistency]
    
You are participating in a competitive game where your 
goal is to identify the most likely abstract source(s)
(e.g., human, instrumental, etc.) that is/are producing
sound in a given audio clip. This clip was broken down
into several frames, each containing multiple audio 
outputs generated by different AIs, representing sounds
at a specific timestamp.

Each frame corresponds to a different moment in the same
video clip and some frames may contain no sound-producing
objects at all, or the text output could provide 
misleading information.

~

Your task:

\textbf{\textcolor{teal}{- Consider the relationships among frames.}}

- Identify and output only the object(s) producing sound
  in each frame.

- For each frame, provide your guess in one line, 
  (seperate by comma if multiple objects), enclosed in 
  with <answer> and </answer> tag pair.
\end{tcolorbox}

\noindent\textbf{User input}:

\begin{tcolorbox}[boxstyle2, title=User Input for Translator with Frame Consistency]
    
<frame0>

     ~wind is blowing and a car engine is running

</frame0>

<frame1>

     ~rain is falling and the wind is blowing

</frame1>

<frame2>

     ~a motor is running and a car engine is revving.

</frame2>

<frame3>

~     a car engine is revving up and revving down.

</frame3>

<frame4>

     ~a car engine is running and a car engine is running.

</frame4>
\end{tcolorbox}

\subsection{Translator with both \textcolor{cyan}{Prompt}-wise and \textcolor{teal}{Frame}-wise consistency}

\textbf{System prompt}:

\begin{tcolorbox}[boxstyle1, title=System Prompt for Translator with Prompt and Frame Consistency]
    
You are participating in a competitive game where your 
goal is to identify the most likely abstract source(s)
(e.g., human, instrumental, etc.) that is/are producing
sound in a given audio clip. This clip was broken down
into several frames, each containing multiple audio 
outputs generated by different AIs, representing sounds
at a specific timestamp.

Each frame corresponds to a different moment in the same
video clip and some frames may contain no sound-producing
objects at all, or the text output could provide 
misleading information.

~

Your task:

\textbf{\textcolor{cyan}{- Analyze the outputs from all audio AIs in each frame together.}}

\textbf{\textcolor{teal}{- Consider the relationships among frames.}}

- Identify and output only the object(s) producing sound
  in each frame.

- For each frame, provide your guess in one line, 
  (seperate by comma if multiple objects), enclosed in 
  with <answer> and </answer> tag pair.
\end{tcolorbox}

\noindent\textbf{User input}:

\begin{tcolorbox}[boxstyle2, title=User Input for Translator with Prompt and Frame Consistency]
    
<frame0>

~<exp1>wind is blowing and a car engine is running </exp1>
 
~<exp2>a stream of water is flowing. </exp2>
 
~<exp3>the sounds of water, wind and wind noise (microphone) </exp3>

</frame0>

<frame1>

~<exp1>rain is falling and the wind is blowing. </exp1>
 
~<exp2>a sound is being recorded. </exp2>

~<exp3>wind </exp3>

</frame1>

<frame2>

~<exp1>a motor is running and a car engine is revving. </exp1>

~<exp2>engine2. sound of an engine </exp2>

~<exp3>engine </exp3>

</frame2>

<frame3>

~ <exp1>a car engine is revving up and revving down. </exp1>
 
 ~<exp2>engine sound. i recorded a sound from a car engine. </exp2>
 
 ~<exp3>engine </exp3>

</frame3>

<frame4>

 ~<exp1>a car engine is running and a car engine is running. </exp1>
 
~ <exp2>a sound is being played. </exp2>
 
 ~<exp3>engine </exp3>

</frame4>
\end{tcolorbox}

\subsection{Translator with model consistency}

\textbf{System prompt}:

\begin{tcolorbox}[boxstyle1, title=System Prompt for Translator with Model Consistency]

You are participating in a competitive game: identify the most likely abstract source(s) (e.g., human, animal, instrumental, mechanical) producing sound in a video clip - based only on textual descriptions.

~

You are given:

~

- Multiple image descriptions (Image 0, Image 1, ...). Each is a frame caption or visual summary generated by a separate agent; they do NOT share information.

~

- Multiple audio descriptions (Audio 0, Audio 1, ...). Each describes what the sound is approximately like (e.g., "sounds like a motorcycle idling") and is generated by a separate agent; they do NOT share information.

~

Your required procedure:

~

- Extract visual evidence: For each image description, identify and list the explicit or clearly implied objects.

~

- Extract acoustic evidence: For each audio description, identify the key acoustic cues.

~

- Within- and cross-modality synthesis.

~

- From all audio agents, compare and consolidate the cues into an overall audio profile. This synthesized audio profile does not need to be a verbatim phrase from the given descriptions; it should capture the best generalization of the sound.

~

- Final decision: Use the synthesized audio profile and visual profile to decide which objects are most likely producing the sound. 

~

Output:

~

- First give a clear, concise, step-by-step reasoning that references description labels (e.g., "Image 2 shows a lawnmower; Audio 1 describes a low rumble similar to lawnmower idling - supports lawnmower").

~

- After that reasoning, output the final decision on a single line only, listing the object(s) most likely producing the sound separated by commas when necessary.

~

- Enclose the single-line final answer in `<answer>` and `</answer>` tags and place nothing else on that line. 
\end{tcolorbox}

\noindent\textbf{User input}:

\begin{tcolorbox}[boxstyle2, title=User Input for Translator with Model Consistency]

Image agent 0: The image shows a person with long dark hair wearing a white sleeveless top with a black pattern. They are are holding a green parrot on their shoulder. The parrot has has a light-colored beak and is perched calmly on the person's hand. In the background, there is a black metal cage with a white cushion inside, and some other of the room is visible, including a white wall and a dark-colored object that appears to be a piece of furniture or a shelf. The overall setting seems to be indoors, possibly in a living room or a similar space.

~

Audio agent 0: The audio contains a single word spoken by a female voice in a neutral tone. The word is 'yes'.

~

Audio agent 1: a person is walking on a carpet. someone is making a sound. this audio contains sound events: clothing, domestic sounds and home sounds.

~

Audio agent 2: The audio contains a sound event that is being described.
\end{tcolorbox}

\end{document}